\documentclass{bmvc2k}

\newcommand\blfootnote[1]{%
  \begingroup
  \renewcommand\thefootnote{}\footnote{#1}%
  \addtocounter{footnote}{-1}%
  \endgroup
}
\def\thefootnote{*}\footnotetext{ These authors contributed equally to this work}

\usepackage{lipsum}
\usepackage{enumitem}

\title{CHASE: Robust Visual Tracking via Cell-Level Differentiable Neural Architecture Search}

\addauthor{Seyed Mojtaba Marvasti-Zadeh\thefootnote{}$\dagger$}{mojtaba.marvasti@ualberta.ca}{1,2}
\addauthor{Javad Khaghani\thefootnote{}}{khaghani@ualberta.ca}{1}
\addauthor{Li Cheng}{lcheng5@ualberta.ca}{1}
\addauthor{Hossein Ghanei-Yakhdan}{hghaneiy@yazd.ac.ir}{2}
\addauthor{Shohreh Kasaei}{kasaei@sharif.edu}{3}
\addinstitution{
 Vision and Learning Lab, \\
 University of Alberta, \\
 Edmonton, Canada
}
\addinstitution{
 Digital Image \& Video Processing Lab, \\
 Yazd University, \\
 Yazd, Iran
}
\addinstitution{
 Image Processing Lab, \\
 Sharif University of Technology, \\
 Tehran, Iran
}
\runninghead{Marvasti-Zadeh, Khaghani, et al.}{Visual Tracking via Cell-Level NAS}

\definecolor{shadecolor}{rgb}{0.93,0.93,0.93}
\begin{document}
\maketitle
\setlength{\abovedisplayskip}{3pt}
\setlength{\belowdisplayskip}{3pt}
\blfootnote{$\dagger$ Corresponding author}
\begin{abstract}
A strong visual object tracker nowadays relies on its well-crafted modules, which typically consist of manually-designed network architectures to deliver high-quality tracking results. Not surprisingly, the manual design process becomes a particularly challenging barrier, as it demands sufficient prior experience, enormous effort, intuition, and perhaps some good luck. Meanwhile, neural architecture search has gaining grounds in practical applications as a promising method in tackling the issue of automated search of feasible network structures. 
In this work, we propose a novel cell-level differentiable architecture search mechanism with early stopping to automate the network design of the tracking module, aiming to adapt backbone features to the objective of Siamese tracking networks during offline training. Besides, the proposed early stopping strategy avoids over-fitting and performance collapse problems leading to generalization improvement.
The proposed approach is simple, efficient, and with no need to stack a series of modules to construct a network. Our approach is easy to be incorporated into existing trackers, which is empirically validated using different differentiable architecture search-based methods and tracking objectives. Extensive experimental evaluations demonstrate the superior performance of our approach over five commonly-used benchmarks. 
\end{abstract}
\section{Introduction}
\label{sec1:intro}
\textit{Visual object tracking} (VOT) aims to localize an unknown object in sequential video frames, just given its initial state. Visual trackers constantly seek to find more robust and accurate approaches considering various applications and challenges in real-world scenarios. In the spirit of \textit{deep learning} (DL), an important objective is to design reliable network architectures for visual tracking purposes \cite{OurSurvey}, usually requiring adequate experience, insightful knowledge, learning heuristics, and extensive manual trial \& error.
\vskip .1cm
\newpage 
\vskip -.5cm
\indent\textit{Neural architecture search} (NAS) has been developed to automatically discover preferable (or ideally optimal) network architecture for a learning task by exploring a wide-reaching space of operation candidates. Generally, NAS methods are classified into the \textit{reinforcement learning} (RL)-based, \textit{evolutionary algorithm} (EA)-based, \textit{Bayesian optimization} (BO)-based, and gradient-based methods, according to their diversified search strategies. Although the first three categories suffer from less efficiency, high time consumption, and extensive computational overhead, the gradient-based methods provide competitive performances \& quite efficiency. The well-known \textit{differentiable architecture search} (DARTS) \cite{DARTS} introduces a generic approach that relaxes the search space into the continuous domain and shares the parameters among candidate architectures. Although DARTS has achieved promising results by its gradient-based searches (resulting in the 1st- \& 2nd-order versions of DARTS), several works \cite{PDARTS,DARTS+,RobustDARTS,FairDARTS,DARTS-,ProxylessNAS} have been proposed to study and address its problems. 

\begin{figure}[t!]
\centering
\includegraphics[width=.95\linewidth]{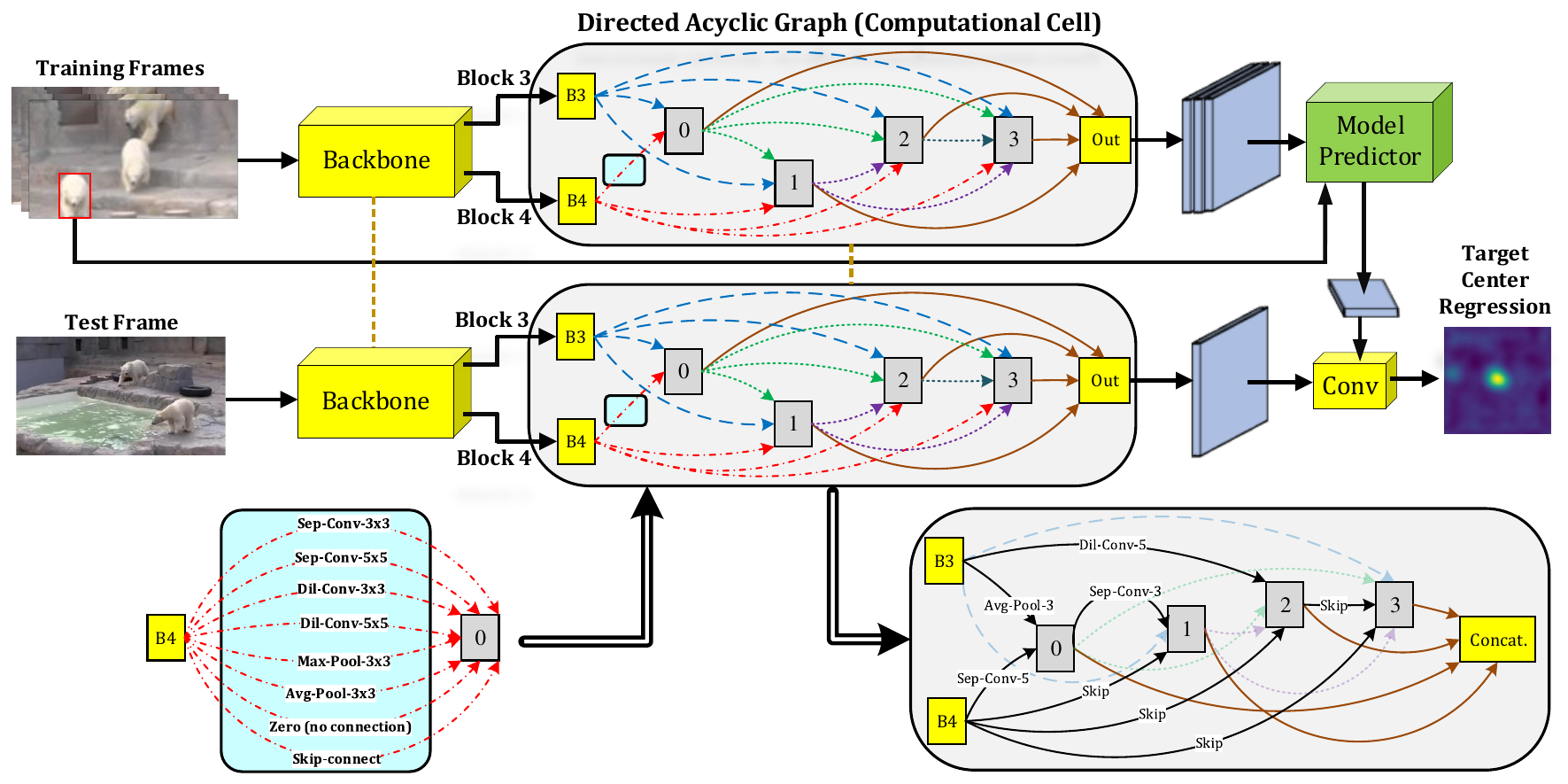}
\vskip -.2cm
\caption{An overview of the proposed CHASE tracker. Cell-level NAS is integrated into the TCR network of the baseline tracker \cite{PrDiMP} to adapt backbone features to the network objective. First, a computational cell is formed in searching phase in which each edge (dash line) is a mixture of candidate operations (shown as a blue box for one edge), each intermediate node is connected to all the previous nodes, and the output node is the concatenation of intermediate nodes (shown by brown solid lines). The objective of this phase is to find the optimal sub-graph (i.e., the best cell shown at the bottom-right) by jointly optimizing the weights and architecture parameters of the cell. Then, in training phase, the computational cell is replaced by the best cell, and the whole pipeline is trained from scratch. Finally, the network is used in evaluating phase for visual tracking.}
\vskip -.4cm
\label{fig:blockdiagram}
\end{figure}

\indent Despite the exploitation of NAS in numerous tasks (e.g., classification \cite{DARTS,FairDARTS}, detection \cite{DetNAS,NAS-FCOS}, semantic segmentation \cite{Auto-DeepLab,Nas-Seg1}), almost all the network architectures for visual tracking are based on human-designed heuristics. Very recently, the LightTrack \cite{LightTrack} uses evolutionary search to obtain lightweight architectures for resource-limited hardware platforms. Also, it uses single-path uniform sampling and lightweight building blocks to achieve more compact architectures and reduce the computational costs. However, single-path sampling decouple the optimizations of the weights and architecture parameters of the supernet, leading to large-variance to the optimization process and tendency to a non-complex structure \cite{NAS-FCOS}. The LightTrack \cite{LightTrack} has inherited the limitations of EA-based methods as well as single-pass search approaches. Furthermore, it searches within a limited search space and stacks the basic blocks to construct the final architecture. \\
\indent In contrast, the aim of this work is to automatically discover the best architecture block (or cell) that adapts large-scale trained backbone features to the objectives of Siamese tracking networks. Although it modifies DARTS \cite{DARTS} with attractive advantages (e.g., weight-sharing \& efficiency), the primary differences include (i) cell-level NAS instead of searching stacked cells together, (ii) integrating cell-level NAS into Siamese framework especially beneficial for visual tracking, (iii) employing operation-level Dropout without hand-crafted constraints used in \cite{PDARTS,PDARTS_IJCV}, and (iv) proposing an early-stopping strategy for searching procedure to address the over-fitting problem and multiple retraining from scratch to select the best cell. The proposed approach (CHASE) takes advantage of the 2nd-order DARTS by learning a cell into Siamese tracking networks. This is contrary to prior works (e.g., \cite{DARTS,PDARTS,PDARTS_IJCV,DARTS+,DARTS-,FairDARTS,RobustDARTS}) searching for multiple stacked cells in CNN/RNN architectures using the simple 1st-order DARTS with lower performance. The CHASE provides a simple, efficient, and generalizable approach considering visual tracking purposes, i.e., high performance \& speed. Besides, DARTS-based methods require searching on a small proxy dataset and transferring the architecture blocks to the large-scale target task to address the high GPU memory consumption issues. However, the CHASE performs a cell-level architecture search, which allows directly utilizing a large-scale tracking dataset. Last but not least, this work removes prior heuristics since the proposed early-stopping provides a performance-aware cell derivation strategy during the searching phase. It exploits a hold-out sample set for validating the generalization of the best cell. Thus, it finds the saturated searching point to address the over-fitting problem and the performance gap between the search and evaluation phases \cite{PDARTS_IJCV}, and then it can select the best cell without requiring multiple retraining from scratch.
Finally, the effectiveness of NAS exploitation and its generalization is validated by employing three versions of DARTS \cite{DARTS,FairDARTS} and integrating the proposed approach into two visual trackers \cite{DiMP,PrDiMP}. \\
\indent In summary, the main contributions are as follows:
\begin{itemize}[noitemsep,nolistsep]
    \item A novel cell-level differentiable architecture search mechanism is proposed to automate the network design of the tracking module during offline training. It is effectively integrated into Siamese tracking network architectures to directly optimize a cell on a large-scale tracking dataset. Our approach is simple, efficient, and easy to be incorporated into existing trackers for improving performance.
    \item An early-stopping strategy is proposed to improve the generalization performance of selected cell architecture. This simple yet effective performance-aware cell derivation strategy finds the best cell during the searching phase without requiring inefficient multiple re-training from scratch.
    \item Extensive experimental evaluations on five widely-used visual tracking benchmarks demonstrate the superior performance of the proposed approach. Moreover, it is practically shown to boost the overall performance when applied to existing baselines. 
\end{itemize}
\section{Related Work}
\label{sec2:relatedwork}
\subsection{Single Object Tracking}
\label{sec:sot}
Most recent state-of-the-art visual trackers are based on classic/custom Siamese networks \cite{SiamRPN++,SiamBAN,SiamCAR,SiamAttn,DiMP,PrDiMP} providing a good trade-off between performance \& computational complexity. The main ideas include taking powerful backbone features and employing lightweight modules to extract robust target-specific features for visual tracking. For instance, all the  SiamRPN++ \cite{SiamRPN++}, SiamBAN \cite{SiamBAN}, SiamCAR \cite{SiamCAR}, DiMP \cite{DiMP}, SiamAttn \cite{SiamAttn}, and PrDiMP \cite{PrDiMP} trackers use ResNet-50 \cite{ResNet} as the backbone and adapt the features for visual object tracking using shallow sub-networks. However, these hand-designed sub-networks are biased toward human priors with no guarantees achieving the highest effectiveness. This motivates this work to automatically design these modules by a cell-level search procedure.
\vskip -.2cm
\subsection{Differentiable NAS}
\label{sec:nas}
Recently, the gradient-based NAS has shown promising results while searching for a few GPU days. As mentioned before, DARTS \cite{DARTS} is the most popular gradient-based approach introducing the 1st- \& 2nd-order approximation-based approaches according to the calculation of architecture gradient, where the 2nd-order one leads to better performance but lower search speed. However, the DARTS suffers from (i) the performance gap between the search \& evaluation phases \cite{PDARTS,PDARTS_IJCV}, (ii) repeating blocks restriction \cite{ProxylessNAS}, (iii) performance collapse \cite{DARTS+,DARTS-,RobustDARTS} due to the model over-fitting, (iv) degenerate architectures \cite{RobustDARTS}, (v) aggregation of skip connections \cite{FairDARTS,PDARTS,PDARTS_IJCV}, and (vi) requiring multiple re-training from scratch. \\
\indent Consequently, several works are presented to address the problems of DARTS. To bridge the gap between the search and evaluation phases, the \textit{progressive DARTS} (PDARTS) \cite{PDARTS} gradually increases the network depth assisted by the search space approximation and regularization. The ProxylessNAS \cite{ProxylessNAS} proposes learning architectures on large-scale datasets, path-level pruning, and latency regularization loss to address repeating blocks restriction, GPU memory consumption, and hardware limitations. The DARTS+ \cite{DARTS+} proposes an early stopping paradigm with hand-crafted constraints to avoid the performance collapse of DARTS due to the model over-fitting in the search phase. To improve the robustness, the RobustDARTS \cite{RobustDARTS} introduces an adaptive regularization and early stopping criterion with the dominant Hessian eigenvalue of validation loss. The DARTS- \cite{DARTS-} distinguishes two roles of skip connections (i.e., stabilization of supernet training \& candidate operation) by an auxiliary skip connection between every two nodes. Finally, the Fair-DARTS \cite{FairDARTS} proposes the collaborative competition approach and auxiliary loss to address the aggregation of skip connections \& discretization discrepancy problems, respectively. \\
\indent Most DARTS-based methods (e.g., \cite{PDARTS,PDARTS_IJCV,DARTS-,DARTS+,RobustDARTS,FairDARTS}) employ the 1st-order DARTS to reduce computational complexity, allowing the search procedure on some stacked cells. The 2st-order DARTS fully exploits training \& validation information and converging to a better local optimum. This work integrates a modified cell-level 2nd-order DARTS into the Siamese framework to track visual targets. The proposed early-stopping strategy and operation-level Dropout \cite{PDARTS,PDARTS_IJCV} without any constraints are exploited to address the over-fitting problem, test-validation performance gap, and the best cell architecture selection.
\section{Proposed Approach: CHASE}
\label{sec3:ours}
The primary motivation is to automatically adapt the robust features extracted from the backbone to the tracking objective by a computational cell (see Fig.~\ref{fig:blockdiagram}). Hence, this work exploits a modified version of DARTS \cite{DARTS} that forms an ordered \textit{directed acyclic graph} (DAG) with $\mathcal{N}$ nodes as its computational cell, which is learned through architecture search procedure. 
The CHASE learns a cell integrated into a Siamese tracking architecture to avoid dramatically affecting the computational complexity \& tracking speed.
PrDiMP \cite{PrDiMP} is used as the baseline to demonstrate the effectiveness of the proposed approach for visual tracking. It includes the \textit{target center regression} (TCR) \& \textit{bounding box regression} (BBR) networks, while it predicts the conditional probability density to minimize the \textit{Kullback-Leiber} (KL) divergence between the predictions and label distribution (see \cite{PrDiMP} for more details). The CHASE tracker replaces additional convolutional blocks after the backbone with a DAG to find the best operations and node connections. 
\subsection{Cell-Level NAS for Visual Tracking}
\label{subsec:cell-darts}
\indent In this section, DARTS is adapted to a Siamese tracking network to move toward our objectives and critical aspects of visual tracking.
In proposed approach, the computational cell has two input nodes and four intermediate nodes. The CHASE fuses multi-level deep features extracted from \texttt{Block3} \& \texttt{Block4} of ResNet-50 \cite{ResNet} in designing the cell, according to their importance for visual tracking \cite{OurSurvey,SurveyDeepTracking}. Given a feature map $\mathcal{X}^{(i)}$ at node $i$, the corresponding latent representation at intermediate node $j$ is computed as $\mathcal{X}^{(j)}=\sum_{i<j}\mathfrak{p}^{(i,j)}(\mathcal{X}^{(i)})$,
where $\mathfrak{p}^{(i,j)}$ stands for candidate operations (from a predefined set $\mathcal{P}=\{\mathfrak{p}_{1}^{(i,j)}, \mathfrak{p}_{2}^{(i,j)}, ..., \mathfrak{p}_{\mathcal{M}}^{(i,j)}\}$ in the search space) on edge $\zeta^{(i,j)}$. Since the DARTS tends to aggregate skip connections due to the rapid error decay during its optimization \cite{FairDARTS,DARTS+}, the CHASE employs the operation-level Dropout without constraints in \cite{PDARTS,PDARTS_IJCV} with an initial rate $\tau$, which gradually decays during the search procedure. The CHASE does not control the number of skip connections to preserve flexibility in cell design and improve training stability. To relax the problem into a continuous search space, the mixed output for $\zeta^{(i,j)}$ is calculated by
\begin{gather} \label{eq:mixed}
\bar{\mathfrak{p}}^{(i,j)}(\mathcal{X})=\sum_{\mathfrak{p} \in  \mathcal{P}}\frac{exp (\alpha_{\mathfrak{p}}^{(i,j)})}{\sum_{{\hat{\mathfrak{p}}}\in \mathcal{P}}exp(\alpha_{{\hat{\mathfrak{p}}}}^{(i,j)})} \mathfrak{p}(\mathcal{X}),
\end{gather} 
in which $\alpha_{\mathfrak{p}}^{(i,j)}$ is the operation mixing weight associated with the operation $\mathfrak{p}$ between nodes $i$ and $j$. By doing so, the cell architecture search converts into the learning of parameters $\alpha = \{\alpha_1^{(i,j)}, \alpha_2^{(i,j)},...,\alpha_{\mathcal{M}}^{(i,j)}\}$. To jointly learn network parameters ($\mathcal{W}$) and architecture parameters ($\alpha$), the \textit{gradient descent} (GD) algorithm is used to minimize the training ($\mathcal{L}_{tr}$) and validation losses ($\mathcal{L}_{val}$) by performing the bi-level optimization problem
\begin{gather} \label{eq:darts_loss}
\min\limits_{\alpha}\;\;\mathcal{L}_{val}(\mathcal{W}^*(\alpha),\alpha) \\
\textnormal{s.t.}\;\;\; \mathcal{W}^*(\alpha)=\argmin\limits_{\mathcal{W}}\mathcal{L}_{tr}(\mathcal{W},\alpha). 
\end{gather} 
To avoid expensive inner optimization, the DARTS reduces the evaluation of architecture gradient by applying the finite difference approximation. By doing so, the 2nd-order approximation of DARTS requires two forward passes for $\mathcal{W}$ and two backward passes for $\alpha$, contrary to the 1st-order DARTS requiring one forward pass for each one (see \cite{DARTS} for more details). \\
\indent The 1st-order DARTS provides the ability to search an architecture by stacking multiple cells according to its simplicity and low complexity, e.g., \cite{FairDARTS,PDARTS,PDARTS_IJCV,RobustDARTS,DARTS+,DARTS-}. Although differentiable NAS aims at minimizing the validation loss to find optimal architectures, the 1st-order DARTS cannot guarantee that the validation loss is sufficiently small due to ignoring the optimization on fully-trained weights $\mathcal{W}^*(\alpha)$. The 2nd-order DARTS embeds the training loss in updating architecture parameters. Hence, it achieves more stability and higher performance than the 1st-order DARTS by fully exploiting training \& validation information and converging to a better local optimum. However, it increases the computational complexity not efficient for optimizing stacked cells. The CHASE enjoys the modified 2nd-order DARTS according to learning one cell that adapts large-scale trained backbone features to the tracking objectives. 
Moreover, the DARTS \cite{DARTS} suffers from some problems as i) deriving the best discrete architecture with the best validation performance by re-training top-$k$ architectures ($k=4$) from scratch, and ii) the performance collapse and over-fitting problems on the validation set, resulting in poor generalization on test datasets. To address these challenges, the proposed CHASE focuses on cell-level search and proposes an early stopping strategy to address the over-fitting problem and multiple re-training from scratch.
\subsection{Early Stopping}
\label{subsec:earlystop}
To alleviate the test-validation gap of DARTS, prior works (e.g., \cite{DARTS+,RobustDARTS}) impose strong early stopping priors or extra computing costs. However, these methods run several times and re-train each best architecture from scratch to select the final one. This work performs a performance-aware cell derivation by the proposed early stopping strategy to address these limitations simultaneously. In particular, generic visual tracking seeks to learn target models generalizable to various appearance changes and real-world challenging scenarios. Hence, the proposed strategy introduces a hold-out sample set represented for generalization validation. 
Note that the CHASE never uses test sets for this purpose. While the CHASE respectively optimizes $\mathcal{W}$ and $\alpha$ on the training and validation sets, it calculates the hold-out loss ($\mathcal{L}_{ho}$) of mixture operations. Then, it derives the best cell architecture at the minimum hold-out loss on the hold-out set by $\mathfrak{p}_{o}^{(i,j)}=\argmax_{\mathfrak{p} \in \mathcal{P}}\alpha_{\mathfrak{p}}^{(i,j)}$. This search-stage cell selection originates from the reduced discrepancies between the continuous cell encoding and the derived discrete cell due to the searching one cell using the proposed modified 2nd-order DARTS, resulting in no several re-training requirements from scratch. That is, the CHASE finds the best cell during the searching phase and then trains it from scratch once.
\section{Empirical Experiments}
\label{sec4:exp}
Herein, the implementation details of the proposed approach, ablation analysis, and tracking results of the best cell architecture on benchmark datasets are reported. Also, codes \& experimental results are publicly available on \href{https://github.com/VisualTrackingVLL}{github.com/VisualTrackingVLL}.
\subsection{Implementation Details}
\label{sec:imp_details}
The backbone consists of ResNet-50 architecture \cite{ResNet} initialized with the pre-trained Image-Net \cite{ImageNet} weights. The offline experiments comprise the searching and training phases. The proposed CHASE tracker is implemented in PyTorch and runs $23$ \textit{fps} on a single Nvidia Tesla V100 GPU with 16GB RAM. Except for the following details, the rest of the hyper-parameters are set to the ones in \cite{PrDiMP}. The test sets are never utilized in searching or training phases.

\subsubsection{Searching Phase}
\label{sec:search}
In this phase, the cell architecture is searched by the modified 2nd-order DARTS. The cell includes $14$ edges and $7$ nodes ($2$ input, $4$ intermediate, and $1$ output), which the output node is obtained by depthwise concatenation of intermediate nodes. The standard DARTS search space is employed to exploit the maximum number of nodes \& edges allowing in a cell, which provides the highest flexibility in cell design. The candidate operations include 3$\times$3 \& 5$\times$5 separable convolutions, 3$\times$3 \& 5$\times$5 dilated convolutions, 3$\times$3 max pooling, 3$\times$3 average pooling, zero (no connection), and skip connection (i.e., $\mathcal{M}=8$).
The CHASE applies operation-level Dropout, which its rate starts from $\tau=0.6$ and gradually decayed to the last epoch. In contrast to \cite{PDARTS,PDARTS_IJCV}, the CHASE fairly explores all operations, considering the importance of skip-connections on the evaluation accuracy and architecture stability. \\
\indent The training set of the TrackingNet dataset \cite{TrackingNet} is divided into two subsets for optimizing the weights of network ($\mathcal{W}$) \& encoding weights of architecture ($\alpha$) on the training ($\mathcal{L}_{tr}$) \& validation ($\mathcal{L}_{val}$) sets, respectively. Besides, the training sets of GOT-10k \cite{GOT-10k} and LaSOT \cite{LaSOT} datasets are used as the hold-out set ($\mathcal{L}_{ho}$) to specify the best architecture among three runs (with different random seeds) and select the final cell architecture based on their performance. Based on the training tricks of NAS in \cite{AutoFPN}, the backbone and BBR parameters are frozen during architecture search, while the architecture parameters are started to optimize after $10$ epochs. It is more critical for the proposed approach to calculate reliable 2nd-order gradients of architecture parameters built on 1st-order ones of network weights. The proposed CHASE provides better initialization of candidate operations directly impacting the optimization procedure of architecture parameters. Thus, it provides fair competition between weight-free operations with other ones and helps effective learning of architecture parameters, leading to performance improvement, acceleration, and avoiding getting stuck into bad local optima.
The network is trained for at most $70$ epochs with a batch size of $10$, similar to the baseline \cite{PrDiMP}. However, the proposed approach stops the training procedure based on the proposed early-stopping strategy (epoch $41$ for CHASE). The Adam optimizer \cite{ADAM} is used to learn network and architecture parameters. The initial learning rate is $0.001$ for optimizing $\mathcal{W}$ with the cosine annealing scheduler. The maximum iteration numbers are $15$K, $15$K, and $5$K for training, validation, and early-stopping procedures. The search phase takes about 41 (18) hours for the second (first) order DARTS method using the TrackingNet dataset on a Nvidia Tesla V100 GPU with 16GB RAM.

\subsubsection{Training Phase}
\label{sec:train}
In contrast to prior works (e.g., \cite{DARTS,PDARTS,PDARTS_IJCV,DARTS+,DARTS-,RobustDARTS}), the CHASE just trains the best model selected in searching phase from scratch.
In this phase, computational cell is replaced by the best cell architecture, and the whole network (including backbone, TCR, and BBR) is jointly trained from scratch for $70$ epochs. The TCR and BBR layers are initialized with random weights ignoring the weights during the searching phase. For the training phase, the training sets of LaSOT \cite{LaSOT}, TrackingNet \cite{TrackingNet}, GOT-10k \cite{GOT-10k}, and COCO \cite{MSCOCO} datasets are used, similar to the baseline \cite{PrDiMP}. Also, other hyper-parameters are set as in the baseline tracker \cite{PrDiMP}.

\subsubsection{Evaluating Phase}
\label{sec:eval}
After offline training phases, the proposed CHASE tracker is evaluated on test splits of generic and aerial visual tracking datasets, namely GOT-10k \cite{GOT-10k}, TrackingNet \cite{TrackingNet}, LaSOT \cite{LaSOT}, UAV-123 \cite{UAV123}, and VisDrone-2019-test-dev \cite{VisDrone2019}. In the online phase, all procedures and settings are the same as \cite{PrDiMP}.

\begin{figure}[b!]
\vskip -.3cm
\centering
\includegraphics[width=1\linewidth]{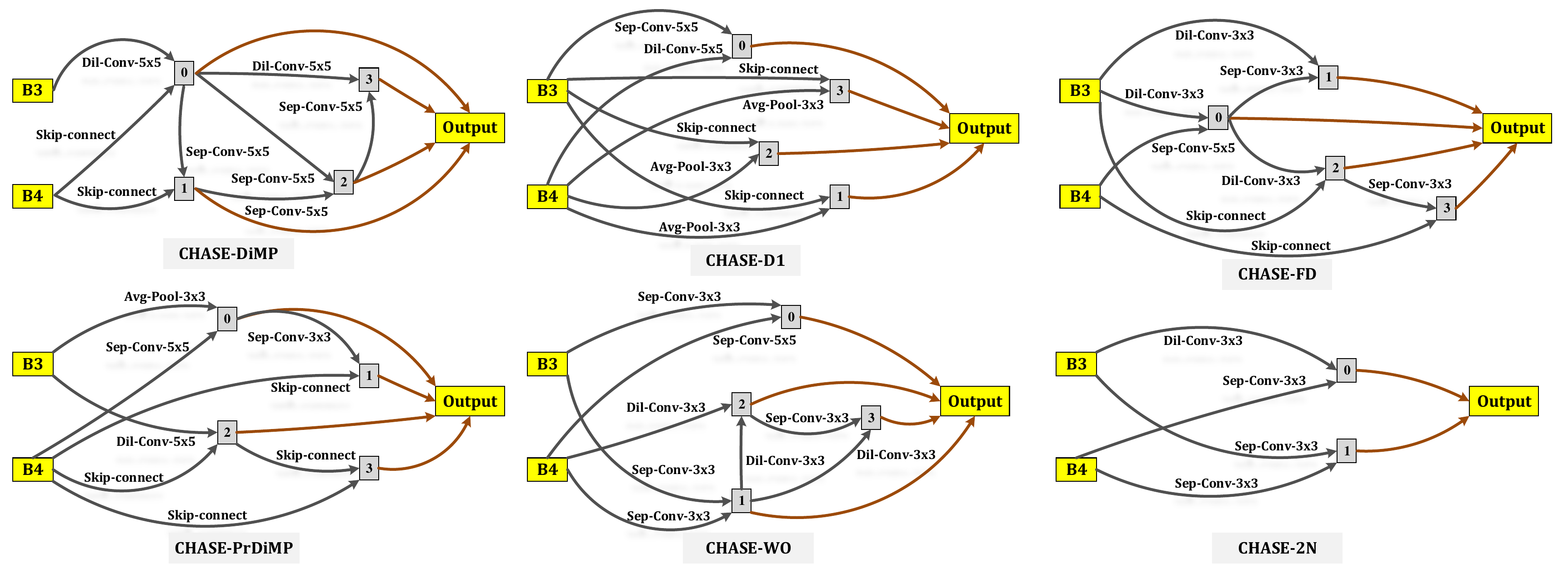}
\vskip -.3cm
\caption{Best cell architectures derived by CHASE-DiMP (modified 2nd-order DARTS), CHASE-D1 (1st-order DARTS), CHASE-FD (Fair-DARTS), CHASE-PrDiMP (modified 2nd-order DARTS), CHASE-WO (modified 2nd-order DARTS without weightless operations), and CHASE-2N (modified 2nd-order DARTS with two intermediate nodes).
\colorbox{yellow}{B3} and \colorbox{yellow}{B4} are the input latent representations (from \texttt{Block3} \& \texttt{Block4} of Resnet50 \cite{ResNet}, respectively). Also, \colorbox{shadecolor}{0}, \colorbox{shadecolor}{1}, \colorbox{shadecolor}{2},
\colorbox{shadecolor}{3} are the intermediate nodes, and the \colorbox{yellow}{output} is the depthwise concatenation of intermediate nodes.}
\label{fig:cells}
\end{figure}

\subsection{Ablation Analysis}
\label{sec:ablation}
In this section, a systematic ablation analysis on the GOT-10k dataset \cite{GOT-10k} is conducted to validate the effectiveness of various search spaces and methods. It includes the cells derived by the 1) 1st-order DARTS (CHASE-D1), 2) Fair-DARTS \cite{FairDARTS} (CHASE-FD), and 3) proposed approach (CHASE-PrDiMP or CHASE). Besides, the CHASE is integrated into the DiMP tracker \cite{DiMP} (CHASE-DiMP), demonstrating the generalization of the proposed approach for visual tracking. Furthermore, three versions of the proposed approach are investigated, including the CHASE with 1) fully segregated datasets in searching \& training phases (CHASE-S/T), 2) a search space consisting of two intermediate nodes (CHASE-2N), and 3) a search space without weightless candidate operations (CHASE-WO). The comparison results are reported in Table~\ref{Ablation} regarding the derived cells shown in Fig.~\ref{fig:cells}. \\
\indent Accordingly, the CHASE-D1 derives a cell dominated by weight-free operations (i.e., skip and pooling operations), and there is no connection between intermediate nodes resulting in a shallow architecture. The CHASE-FD employs the Fair-DARTS \cite{FairDARTS}, which utilizes the Sigmoid activation function and an auxiliary loss to address exclusive competition of skip-connections and discretization discrepancy. Nonetheless, the CHASE outperforms the CHASE-D1 \& CHASE-FD up to $3.6\%$ and $2.1\%$ in terms of \textit{average overlap} (AO) metric, respectively. 
Conventional DARTS-based methods (with stacked cell networks for image classification) search a network architecture on a small proxy dataset (e.g., CIFAR-10) and then transfer it to a large-scale target dataset (e.g., ImageNet) to alleviate high memory consumption \cite{ProxylessNAS}. However, the proposed approach can enjoy searching on the large-scale TrackingNet dataset by its cell-level search. Hence, the CHASE uses the large-scale TrackingNet dataset in both searching \& training phases outperforming the CHASE-S/T up to $1.4\%$ in terms of AO metric. Except for CHASE-S/T, all CHASE-versions have been searched and trained on similar datasets mentioned in Sec.~\ref{sec:search} and Sec.~\ref{sec:train}, respectively. \\
\indent While the CHASE employs the standard DARTS search space to have more design flexibility via the maximum number of nodes \& edges allowing in a cell, the CHASE-2N and CHASE-WO represent search spaces with limited node numbers (i.e., two intermediate nodes) and removed weightless candidate operations (i.e., pooling, zero, \& skip connect), respectively. According to the results, the CHASE has improved the performance of CHASE-2N \& CHASE-WO up to $2.8\%$ \& $6.3\%$ in terms of AO metric, respectively. These results demonstrate prior heuristics and limited search space dramatically affect architecture design and tracking performance. For instance, the intuitive reason in the case of CHASE-WO is that removing weightless operations (particularly skip-connections) has been led to instability in cell design and accuracy degradation. Besides, the node restriction results in shallow cell architecture and limited performance improvement.
The computational cells derived by the CHASE-PrDiMP confirm selecting various operations regarding objective function, increasing the depth as necessary, and preventing over-fitting and performance collapse problems.
Finally, the proposed approach is integrated into the DiMP tracker \cite{DiMP} minimizing an L\textsuperscript{2}-based discriminative learning loss to train its network to investigate the generalization to different objective functions. The proposed approach outperforms the DiMP tracker \cite{DiMP} up to $2.5\%$ in terms of the AO and up to $3.6\%$ in terms of \textit{success rate} (SR) at the overlap threshold of $0.5$. At last, the best-performing tracker, CHASE, is selected to be compared with recent trackers in the next section.

\begin{table}[t!]
\scriptsize
\caption{Ablation analysis of CHASE on GOT-10k dataset \cite{GOT-10k}.} 
\centering 
\resizebox{\textwidth}{!}{
\begin{tabular}{c | c c | c c c c c c c} 
\hline \hline
Metric & DiMP \cite{DiMP} & \textbf{CHASE-DiMP} & PrDiMP \cite{PrDiMP} & CHASE-D1 & CHASE-FD & \textbf{CHASE} & CHASE-2N & CHASE-WO & CHASE-S/T  \\ \hline \hline
SR\textsubscript{$0.75$} ($\uparrow$) & 49.2 & \textcolor{red}{\textbf{51.1}} & 54.3 & 54.8 & 56.1 & \textcolor{red}{\textbf{56.5}} & 51.4 & 45.9 & 56.1 \\ 
SR\textsubscript{$0.5$} ($\uparrow$) & 71.7 & \textcolor{red}{\textbf{75.3}} & 73.8 & 76.7 & 76.8 & \textcolor{red}{\textbf{78.8}}  & 76.5 & 71.5 & 76.3 \\ 
AO ($\uparrow$) & 61.1 & \textcolor{red}{\textbf{63.6}} &  63.4 & 64.9 & 65.6 & \textcolor{red}{\textbf{67.0}}  & 64.2 & 60.7 & 65.6 \\ 
\hline
\end{tabular}
\label{Ablation}}
\vskip -.3cm
\end{table}

\subsection{State-of-the-art Comparison}
\label{sec:comparison}
In this section, the state-of-the-art evaluations are performed on five large-scale visual tracking benchmarks (refer to Sec.~\ref{sec:eval}) and the proposed CHASE tracker is compared with various state-of-the-art visual trackers, namely ECO \cite{ECO}, SiamMask \cite{SiamMask}, DaSiamRPN \cite{DaSiamRPN}, SiamRPN++ \cite{SiamRPN++}, ATOM \cite{ATOM}, DCFST \cite{DCFST}, COMET \cite{COMET}, SiamFC++ \cite{SiamFCpp}, DiMP-50 \cite{DiMP}, PrDiMP-50 \cite{PrDiMP}, KYS \cite{KYS}, SiamAttn \cite{SiamAttn}, MAML \cite{MAML}, ROAM++ \cite{ROAM}, SiamCAR \cite{SiamCAR}, SiamBAN \cite{SiamBAN}, D3S \cite{D3S}, Ocean \cite{Ocean}, and LightTrack \cite{LightTrack}. \\
\noindent\textbf{GOT-10k \cite{GOT-10k}}: This large high-diversity dataset includes over $10$K videos as the training set and $180$ videos for evaluation without publicly available ground-truth. Notably, the target classes for evaluation do not overlap with training ones. Hence, this dataset is usually used for studying the transferability of proposed approaches for tracking unseen targets.  Therefore, the proposed CHASE uses its training set as one of the hold-out sets to early-stop the cell searching phase. The comparison results presented in Table~\ref{tab_results} show that the CHASE outperforms the baseline up to $3.6\%$, $5\%$, and $2.2\%$ in terms of AO and SR at overlap thresholds of $0.5$ and $0.75$, respectively. Besides, the CHASE has achieved better results ($4.7\%$ in AO, $6.2\%$ in SR\textsubscript{0.5}) compared with the LightTrack \cite{LightTrack}.\\
\noindent\textbf{LaSOT \cite{LaSOT}}: LaSOT is a long-term and challenging tracking benchmark consisting of $1400$ videos and $3.5$M frames, with $2500$ frames per video on average. The test set contains $280$ videos and $690$K frames with target disappear/reappear scenarios. Thus, this dataset appropriately indicates the robustness of short-term trackers in real-world situations. For this reason, the proposed tracker uses its training set as the second dataset of hold-out set in the searching phase. As shown in Table~\ref{tab_results}, the CHASE improves the baseline results \cite{PrDiMP} by a margin of $1.9\%$, $2.3\%$, and $2.1\%$ in terms of \textit{area under curve} (AUC), normalized precision, and precision, respectively. \\
\noindent\textbf{TrackingNet \cite{TrackingNet}}: TrackingNet is a challenging in-the-wild tracking dataset consisting of $27$ classes of targets from YouTube videos. This dataset contains more than $30$K videos and $14.4$M frames, including $500$ videos for testing which the ground-truths are not publicly available. From Table~\ref{tab_results}, the MAML tracker \cite{Instance_Det_meta} has close results (better in precision metric) compared with the proposed tracker since it employs a modern object detector (i.e., FCOS \cite{FCOS}) and online domain adaptation to enhance discriminating target from non-target regions. However, the proposed CHASE tracker has achieved better results in terms of AUC and normalized precision, and it has improved the baseline results by a margin of $1\%$ in AUC and $1.4\%$ in precision metric. \\
\noindent\textbf{UAV-123 \cite{UAV123}}: UAV-123 is an challenging aerial-view tracking dataset consisting of $123$ videos, $113$K frames, and $9$ classes of targets captured from a low-altitude perspective. According to the results in Table~\ref{tab_results}, the proposed CHASE tracker outperforms the state-of-the-art visual trackers but also the baseline tracker \cite{PrDiMP} up to $1.2\%$ and $0.8\%$ in terms of success and precision rate metrics. \\
\noindent\textbf{VisDrone-2019-test-dev \cite{VisDrone2019}}: VisDrone-2019 also aims to track visual targets captured from aerial-view. It includes $35$ test videos ($112$K frames) from challenging scenarios such as abrupt camera motion, tiny targets, fast view-point change, and day/night conditions. Compared with the baseline \cite{PrDiMP}, the results of the CHASE tracker have improved up to $1.9\%$ in AUC and $2.3\%$ in precision rate. The COMET \cite{COMET} has obtained the best results employing the training set of VisDrone for its offline training and accurately designed modules for small object tracking. 
\begin{table}[t!]
\scriptsize
\caption{State-of-the-art comparison results on GOT-10k \cite{GOT-10k}, LaSOT \cite{LaSOT}, TrackingNet \cite{TrackingNet}, UAV-123 \cite{UAV123}, VisDrone-2019-test-dev \cite{VisDrone2019} datasets.} 
\centering 
\resizebox{\textwidth}{!}{
\begin{tabular}{c | c c c | c c c | c c c | c c | c c} 
\hline \hline
\multirow{2}{*}{Trackers} & \multicolumn{3}{c|}{GOT-10k} & \multicolumn{3}{c|}{LaSOT} & \multicolumn{3}{c|}{TrackingNet} & \multicolumn{2}{c|}{UAV-123} & \multicolumn{2}{c}{VisDrone-2019-test-dev} \\
& AO ($\uparrow$) & SR\textsubscript{0.5} ($\uparrow$) & SR\textsubscript{0.75} ($\uparrow$) & AUC ($\uparrow$) & Norm. Prec. ($\uparrow$) & Prec. ($\uparrow$) & AUC ($\uparrow$) & Norm. Prec. ($\uparrow$) & Prec. ($\uparrow$) & SR\textsubscript{0.5} ($\uparrow$) & Prec. ($\uparrow$) & AUC ($\uparrow$) & Prec. ($\uparrow$) \\\hline 
\textbf{CHASE} & \textcolor{red}{\textbf{67.0}} & \textcolor{red}{\textbf{78.8}} & \textcolor{red}{\textbf{56.5}} & \textcolor{red}{\textbf{61.7}} & \textcolor{red}{\textbf{71.1}} & \textcolor{red}{\textbf{62.9}} & \textcolor{red}{\textbf{76.8}} & \textcolor{red}{\textbf{82.5}} & \textcolor{blue}{\textbf{71.8}} & \textcolor{red}{\textbf{83.9}} & \textcolor{red}{\textbf{88.2}} & \textcolor{blue}{\textbf{61.7}} & 82.0 \\\hline
LightTrack \cite{LightTrack} & 62.3 & 72.6 & - & - & - & 56.1 & 73.3 & 78.9 & 70.8 & - & - & - & - \\\hline
PrDiMP-50 \cite{PrDiMP} & 63.4 & \textcolor{blue}{\textbf{73.8}} & \textcolor{blue}{\textbf{54.3}} & \textcolor{blue}{\textbf{59.8}} & \textcolor{blue}{\textbf{68.8}} & \textcolor{blue}{\textbf{60.8}} & \textcolor{blue}{\textbf{75.8}} & 81.6 & 70.4 & \textcolor{blue}{\textbf{82.7}} & \textcolor{blue}{\textbf{87.4}} & 59.8 & 79.7 \\\hline
Ocean \cite{Ocean} & 61.1 & 72.1 & 47.3 & 56.0 & 65.1 & 56.6 & - & - & - & - & - & 59.4 & 82.3 \\\hline
D3S \cite{D3S} & 59.7 & 67.6 & 46.2 & - & - & - & 72.8 & 76.8 & 66.4 & - & - & - & - \\\hline
ROAM++ \cite{ROAM} & 46.5 & 53.2 & 23.6 & 44.7 & - & 44.5 & 67.0 & 75.4 & 62.3 & - & - & - & - \\\hline
SiamAttn \cite{SiamAttn} & - & - & - & 56.0 & 64.8 & - & 75.2 & \textcolor{blue}{\textbf{81.7}} & - & 79.4 & 84.5 & - & - \\\hline
KYS \cite{KYS} & 63.6 & 75.1 & 51.5 & 55.4 & 63.3 & - & 74.0 & 80.0 & 68.8 & - & - & - & - \\ \hline
DiMP-50 \cite{DiMP} & 61.1 & 71.7 & 49.2 & 56.9 & 65.0 & 56.7 & 74.0 & 80.1 & 68.7 & 80.4 & 85.5 & 60.8 & 80.5 \\ \hline
SiamCAR \cite{SiamCAR} & 56.9 & 67.0 & 41.5 & 50.7 & 60.0 & 51.0 & - & - & - & 77.3 & 81.3 & - & - \\ \hline
SiamBAN \cite{SiamBAN} & - & - & - & 51.4 & 59.8 & 52.1 & - & - & - & 77.4 & 83.3 & - & -  \\ \hline
MAML \cite{Instance_Det_meta} & - & - & - & 52.3 & - & - & 75.7 & 82.2 & \textcolor{red}{\textbf{72.5}} & - & - & - & - \\ \hline
ATOM \cite{ATOM} & 55.6 & 63.4 & 40.2 & 51.5 & 57.6 & 50.5 & 70.3 & 77.1 & 64.8 & 78.9 & 85.6 & 57.1 & 76.7 \\ \hline
SiamRPN++ \cite{SiamRPN++} & 51.8 & 61.8 & 32.5 & 49.6 & 56.9 & - & 73.3 & 80.0 & 69.4 & 78.8 & 84.0 & 59.9 & 79.1 \\ \hline
DCFST \cite{DCFST} & \textcolor{blue}{\textbf{63.8}} & 75.3 & 49.8 & - & - & - & 75.2 & 80.9 & 70.0 & - & - & - & - \\ \hline
COMET \cite{COMET} & 59.6 & 70.6 & 44.9 & 54.2 & - & - & - & - & - & 79.4 & 86.1 & \textcolor{red}{\textbf{64.5}} & \textcolor{red}{\textbf{83.9}} \\ \hline
SiamFC++ \cite{SiamFCpp} & 59.5 & 69.5 & 47.9 & 54.4 & 62.3 & 54.7 & 75.4 & 80.0 & 70.5 & - & - & - & - \\ \hline
SiamMask \cite{SiamMask} & 51.4 & 58.7 & 36.6 & - & - & - & 72.5 & 77.8 & 66.4 & - & - & 58.1 & 79.4 \\ \hline
DaSiamRPN \cite{DaSiamRPN} & - & - & - & - & - & - & 63.8 & 73.3 & - & 72.6 & 78.1 & - & - \\ \hline
ECO \cite{ECO} & 31.6 & 30.9 & 11.1 & 32.4 & 33.8 & 30.1 & 55.4 & 61.8 & 49.2 & 63.1 & 74.1 & 55.9 & \textcolor{blue}{\textbf{82.6}} \\\hline
\hline
\end{tabular}
\label{tab_results}}
\vskip -.3cm
\end{table}
\vskip -.3cm
\section{Conclusion}
\label{sec5:conclusion}
A novel cell-level differentiable architecture search mechanism is proposed. To address the inherent limitations of differentiable architecture search, we exploit the second-order DARTS by operation-level dropout without any post-processing and introduce early stopping to mitigate the over-fitting and performance collapse issues. Our approach is simple, efficient, and easy to be integrated into existing visual trackers. Extensive experiments demonstrate the effectiveness of the proposed approach, as well as noticeable performance improvement when working with different existing trackers.\\ \\
\textbf{Acknowledgement:} This research was partly supported by the NSERC Discovery Grant (No. RGPIN-2019-04575) and the UAHJIC Grants.
\bibliography{egbib}
\end{document}